\documentclass{article}

\usepackage{arxiv}

\usepackage{algorithm}
\usepackage{algorithmic}

\usepackage{microtype}
\usepackage{graphicx}
\usepackage{subfigure}
\usepackage{booktabs} 
\usepackage{bm}
\usepackage{hyperref}

\usepackage{amsmath}
\usepackage{amssymb}
\usepackage{mathtools}
\usepackage{amsthm}

\usepackage[capitalize,noabbrev]{cleveref}

\theoremstyle{plain}

\theoremstyle{definition}

\theoremstyle{remark}

\usepackage{epstopdf}
\usepackage{float}
\usepackage[justification=centering]{caption}

\graphicspath{ {./fig/} }

\renewcommand{\shorttitle}{\textit{arXiv} Template}

\title{Extreme precipitation forecasting using attention augmented convolutions}
\date{Jan 27, 2022}

\author{ \hspace{1mm}Weichen Huang\\
				St. Andrews College\\
				Dublin, A94 XN72 Ireland \\
		\texttt{w.huang@students.st-andrews.ie} \\
}

\renewcommand{\undertitle}{Technical Report}
\renewcommand{\shorttitle}{Extreme Precipitation Forecasting}

\begin{document}
\maketitle

\begin{abstract}

Extreme precipitation wreaks havoc throughout the world, causing billions of dollars in damage and uprooting communities, ecosystems, and economies. Accurate extreme precipitation prediction allows more time for preparation and disaster risk management for such extreme events. In this paper, we focus on short-term extreme precipitation forecasting (up to a 12-hour ahead-of-time prediction) from a sequence of sea level pressure and zonal wind anomalies. Although existing machine learning approaches have shown promising results, the associated model and climate uncertainties may reduce their reliability. To address this issue, we propose a self-attention augmented convolution mechanism for extreme precipitation forecasting, systematically combining attention scores with traditional convolutions to enrich feature data and reduce the expected errors of the results. The proposed network architecture is further fused with a highway neural network layer to gain the benefits of unimpeded information flow across several layers. Our experimental results show that the framework outperforms classical convolutional models by 12\%. The proposed method increases machine learning as a tool for gaining insights into the physical causes of changing extremes, lowering uncertainty in future forecasts.

\end{abstract}


\section{Introduction}
\label{introduction}

Climate change is one of the most critical issues affecting humanity's future, and it will only grow in importance as extreme weather events increase in frequency and intensity. Among the most recent examples is the European floods that killed over 242 people and cost over 10 billion euros in damage in parts of Germany, France, Switzerland, Italy, and the United Kingdom. Total damages from Atlantic hurricanes in the United States are expected to total \$67 billion in 2021. Hurricane Ida struck the Louisiana coast in August 2021, causing \$64.5 billion in damage and killing 96 people. As climate change intensifies, these natural disasters will become more prevalent and destructive \cite{Ornes2018, Santer2022}.

To date, long-term weather and climate prediction models such as Extreme Value Analysis (EVA) and other numerical weather prediction (NWP) models are becoming increasingly inaccurate and unstable due to the unpredictability of extreme weather events as a direct result of climate change \cite{Seneviratne2018, Santer2022}. It is in our great interest to provide an accurate and reliable system to predict and classify extreme weather events and estimate the damage caused.

Machine learning for climate forecasting \cite{Jones2017, Sijie2020, Gibson2021} has risen in popularity as a consequence of recent developments in machine learning, such as deep learning \cite{shi2015convolutional, yujie2016, Chattopadhyay2020, Davenport21, civitarese2021extreme}, as well as the ability of some machine learning techniques to outperform numerical models in specific tasks \cite{Gibson2021}. A widely applied neural network architecture for this task is the Convolutional Neural Network (CNN) \cite{shi2015convolutional, yujie2016, Davenport21}. Convolutional layers in CNNs learn convolutional filters of size $K \cdot K$, with input and output dimensions $D_{in}$ and $D_{out}$, respectively. The layer is parametrized by a 4D kernel tensor $\bm{W}$ of dimension $K \cdot K \cdot D_{in} \cdot D_{out}$ and a bias vector $\bm{b}$ of dimension $D_{out}$. As a result, CNNs are not spatially invariant to the input data and do not encode the position of each value in the input matrix.

In this paper, we aim at improving the accuracy and applicability of extreme precipitation forecasting. By augmenting the classical convolutional layer with a multi-head self-attention module, we have solved a major drawback with convolutional neural networks, as the self-attention's receptive field is always the full input matrix. This significantly reduces the expected errors in the final results. We demonstrate its applicability to extreme precipitation forecasting using real-world datasets. Furthermore, we propose a novel self-attention augmented convolutional layer that allows us to gain the benefits of the attention network on multi-dimensional feature maps.

\section{Related Work}

\subsection{Convolutional precipitation forecasting}

Convolutional networks have shown to be particularly effective in computer vision applications. According to \cite{Davenport21}, such networks have been used to examine large-scale "extreme precipitation circulation patterns" (EPCP). A traditional CNN classifier was developed to distinguish between EPCP and non-EPCP days. The CNN model accurately identifies 91\% of severe precipitation days in the US midwest as EPCPs, with an overall accuracy of 88\% across both classes. This study further calculated trends in EPCP frequency and precipitation intensity on days when EPCPs occur based on the classification results.

Other network designs, such as Convolutional LSTM Network \cite{shi2015convolutional} and Temporal Fusion Transformer (TFT) \cite{civitarese2021extreme}, have been used to tackle comparable precipitation forecasting issues in addition to traditional convolutional networks.

These efforts, particularly \cite{Davenport21}, influenced our research and experimentation. However, our work puts a greater emphasis on developing more effective and efficient network topologies to increase prediction accuracy and deep learning model performance.

\subsection{Attention mechanisms}

Attention accentuates the relevant features of the input data while omitting the less relevant ones, with the concept that the network should spend more computational resources on a smaller but critical portion of the data \cite{lindsay2021}. Self-Attention, also known as intra attention, connects distinct points of a single sequence to compute a representation of the same sequence \cite{vaswani2017attention}. The self-attention mechanism allows the inputs to interact with one another (self) and determine who they should focus on more (attention). These interactions and attention scores are aggregated in the outputs. Self-attention, as compared to other attention methods, minimizes the total computing complexity of each layer of calculations by lowering the number of consecutive operations required. It may also reduce the maximum path length between any two input and output locations in a network with multiple layer types.

Using self-attention together with convolutions is shared by recent work in Natural Language Processing \cite{yu2018qanet} and Reinforcement Learning \cite{yu2018qanet, yang2019convolutional}. The convolution technique has a critical weakness in that it only works on local neighbourhoods, which means it misses out on global data. Self-attention, on the other hand, is capable of capturing long-range interactions that have been mostly applied to sequential modelling problems. In particular, \cite{cordonnier2020relationship} further proves that a multi-head self-attention layer is "at least as expressive as any convolutional layer".

\section{Methods}

This study aims at tackling the problem of predicting and classifying extreme precipitation days by combining state-of-the-art deep learning representations with novel design techniques. To achieve this, we build a self-attention augmented convolutional neural network to predict severe precipitation days using daily SLP and 500-hPa GPH anomalies. Each day's input data is processed via various layers to produce an output categorization of extreme precipitation (Class 1) or non extreme precipitation (Class 0). Our model receives a three-dimensional matrix with dimensions of $15 \cdot 35 \cdot 2$ for each day (i.e., latitude $\cdot$ longitude $\cdot$ 2 input variables). Precipitation data is used to produce ground-truth labels for model training.

In the following sections, we introduce the datasets we used in the experiments and the corresponding data processing methods. Then we formally describe the proposed Attention-Augumented Convolutions network architecture.

\subsection{Weather datasets}

We compute extreme precipitation days across the United States Midwest from 1981 to 2019 using PRISM 4 km daily precipitation (PRISM Climate Group). We calculate the mean precipitation across a rectangular region including the Upper Mississippi Watershed and the eastern section of the Missouri Watershed (37°N to 48°N, 104°W to 86°W) for each day. We determine the 95th percentile (p95) of daily precipitation using this regional daily precipitation time-series, and extreme precipitation days are classified as those that surpass the p95 threshold. The algorithm for this is presented in \ref{alg:precipitation-percentile}. We examine excessive precipitation days throughout all seasons since extreme precipitation can occur throughout the year in the Midwest.

\begin{figure}[ht]
\vskip 0.2in
\begin{center}
\centerline{\includegraphics[width=0.6\columnwidth]{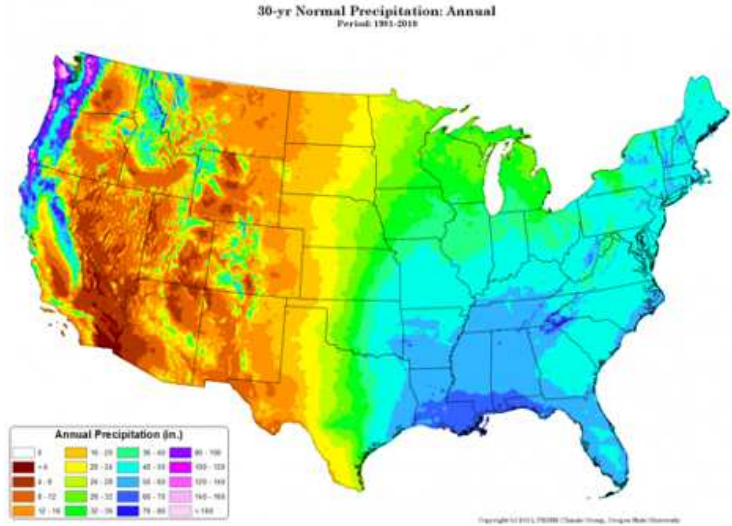}}
\caption{PRISM Dataset}
\label{climatenets-dataset-prism}
\end{center}
\vskip -0.2in
\end{figure}

\begin{algorithm}[tb]
    \caption{Precipitation percentile calculation}
    \label{alg:precipitation-percentile}
 \begin{algorithmic}
    \STATE {\bfseries Input:} array $x_i$, percentile $m$
    \STATE {\bfseries Sort} array.
    \STATE $k = m * (n-1)$
    \STATE $f = floor(k)$
    \STATE $c = ceil(k)$
    \IF{$f == c$}
    \STATE {\bfseries Return} $x_f$.
    \ENDIF
    \STATE $d_0 = x_f * (c-k)$
    \STATE $d_1 = x_c * (k-f)$
    \STATE {\bfseries Return} $d_0 + d_1$
 \end{algorithmic}
 \end{algorithm}

The NCEP/NCAR-R1 reanalysis, which offers worldwide coverage at $2.5° \cdot 2.5°$ horizontal resolution, is used to determine daily mean sea level pressure (SLP) and 500-hPa geopotential height (GPH) anomalies. We examine atmospheric variables over a broader geographical region that includes the entire United States and nearby waters (20°N to 55°N and 140°W to 55°W). To eliminate uniform thermal dilatation produced by tropospheric warming, we first deduct the area-weighted average 500-hPa GPH trend over the atmospheric domain, maintaining spatially non-uniform changes in 500-hPa GPH that may effect severe precipitation. The daily normalized anomalies (z-scores) are then calculated by subtracting the grid-cell calendar-day mean from the grid-cell calendar-day standard deviation. We then apply a threshold to the z-scores to identify extreme precipitation days.

We use zonal wind ($u$), meridional wind ($v$), and specific humidity ($q$) fields from the NCEP/NCAR-R1 reanalysis to evaluate differences in moisture flux on days with extreme precipitation and non-extreme precipitation patterns.

\begin{equation} \label{zscore}
	z = \frac{X - \mu}{\sigma}
\end{equation}
where:
\begin{description} 
	\item $z$ is the computed z-score.
	\item $X$ is the value of the variable at the grid-cell.
	\item $\mu$ is the grid-cell calendar-day mean.
	\item $\sigma$ is the grid-cell calendar-day standard deviation.
\end{description}

\subsection{Attention augmented convolutions}

We propose concatenating convolutional feature maps with a set of feature maps obtained by self-attention to enhance convolutional operators with this self-attention mechanism. Such an operation is defined below in Formula \ref{aaconv}.

Given an input tensor of shape $(H, W, D)$, which represents the height, width, and the feature depth of the anomalies respectively, we flatten it to a matrix $A(1, H \cdot W \cdot D)$ and perform multihead attention as proposed in the Transformer architecture \cite{vaswani2017attention}. The output of the self-attention mechanism for a single head $h$ can be formulated as below (\ref{attn}):

\begin{equation} \label{attn}
	Attn(X) = softmax(\frac{QK^T}{\sqrt{d_k}})V
\end{equation}
where:
\begin{description} 
	\item $Q = (X * W_q)$
	\item $K = (X * W_k)$ 
	\item $V = (X * W_v)$ 
	\item $W_q$, $W_k$, $W_v$ are the learned linear transformations that map the input $X$ to queries ($Q$), keys ($K$), and values ($V$) respectively.
\end{description}

A single attention score calculation uses $O(A^2 \cdot h)$ space, with the multi-head attention calculation using $O(A^2 \cdot h^2)$ space. This is much more efficient than many other previous attention-based augmentations to the convolutional layer due to omitting relative position embeddings. The multi-head attention calculation is performed on a single head at a time, and the output of the multi-head attention is concatenated to the output of the convolutional layer.

\begin{equation} \label{mha}
	MHA(X) = Concat[Attn_1(X), Attn_2(X), ... ,	Attn_Nh(X)] * W_mh
\end{equation}
where:
\begin{description} 
	\item $W_mh$ is the learned linear transformation that maps the concatenated attention scores to the output of the multi-head attention.
\end{description}

Finally, the multi-head attention scores are concatenated to the output of the convolutional layer:

\begin{equation} \label{aaconv}
	AAConv(X) = Concat[Conv(X), MHA(X)]
\end{equation}
where:
\begin{description} 
	\item $Conv(X)$ is the output of the convolutional layer, and $MHA(X)$ is the output of the multi-head attention.
\end{description}

\subsection{Network architecture description}

\begin{figure*}
\vskip 0.2in
\begin{center}
\centerline{\includegraphics[width=0.6\textwidth]{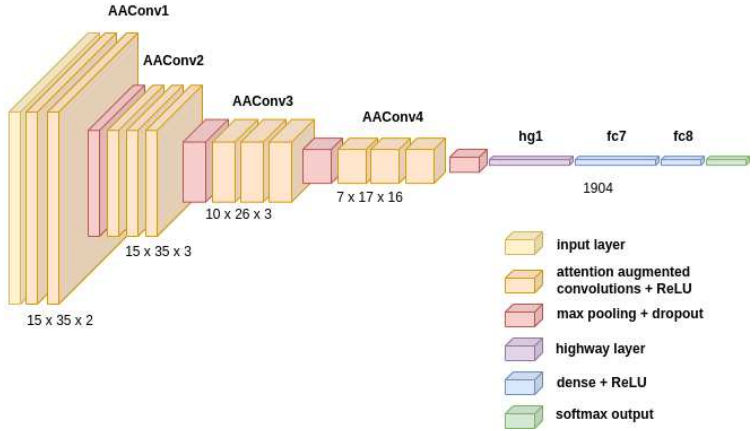}}
\caption{Proposed Network Architecture}
\label{climatenets-network-arch}
\end{center}
\vskip -0.2in
\end{figure*}

The proposed network architecture is shown in Figure \ref{climatenets-network-arch}. Given a multi-channel input tensor formed by stacking the input maps of mean SLP and GPH anomalies, the network predicts if the precipitation of the area is above the p95 threshold. The network comprises two attention augmented convolutional layers, each with a filter of 16, a max-pooling and dropout layer.

A highway layer followed by a dense layer is used to make predictions. Furthermore, while the predictands of our existing model were found to be quite stable, the highway network was added to provide feature depth to the model, thus improving its predictive accuracy.

\subsection{Training and validation}

Each model was trained for 100 epochs with the batch size of 32. The Adam optimizer was used with a constantly decaying learning rate scheduler, with the minimum and maximum learning rates as 10e-4 and 10e-2, respectively. The categorical crossentropy error was implemented as the loss function.

\subsection{Experiments}

In order to evaluate the performance of the proposed models, we implemented the classical CNN model as the baseline. The proposed Self-Attention CNN models are built and trained with two variations: without Highway networks (SAConvNet) and with Highway networks (SAConvNet + Highway). The performance of these two variations is compared with that of
the classical CNN models on exactly the same datasets. 

\begin{figure*}
\vskip 0.2in
\begin{center}
\centerline{\includegraphics[width=\textwidth]{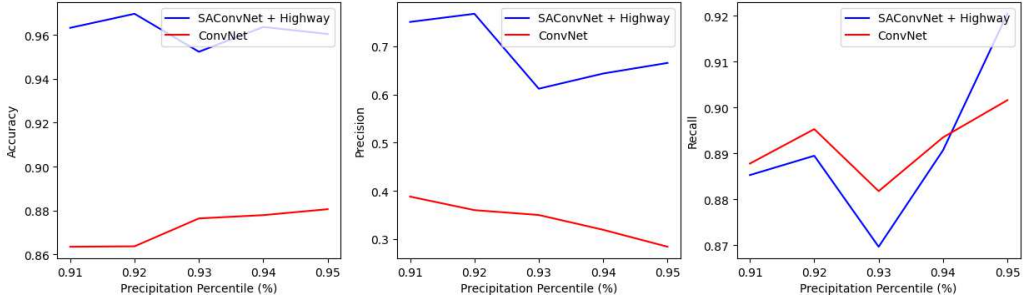}}
\caption{Precipitation Percentile Experiment Metrics}
\label{preciptation-metrics}
\end{center}
\vskip -0.2in
\end{figure*}

To study the model's generalisation ability and robustness, we trained 5 models in each experiment, with identical hyperparameters. Each model was applied to predict the extreme precipitation at 5 given thresholds, namely p91-95. These precipitation thresholds can be used to model real-world scenarios for extreme precipitation. The results of this experiment are shown in Figure \ref{preciptation-metrics}.

Each experiment is repeated with 5 times. The arithmetic Mean of the performance metrics are calculated, together with the Standard Error of the Mean (SEM). This is to remove any random factors that might be introduced by the model training / testing process.

All possible prediction results can be divided into the following four cases:
\begin{itemize}
\item True Positive (TP): actual extreme weather events are classified as extreme.
\item True Negative (TN): actual normal weather events are classified as normal.
\item False Positive (FP): actual normal weather events are classified as extreme events (e.g., false alarms).
\item False Negative (FN): actual extreme weather events are classified as normal.
\end{itemize}

Accordingly, the performance of the proposed models can be evaluated by using the following evaluation metrics:
\begin{itemize}
\item \textbf{Accuracy} measures the proportion of the correctly classified extreme weather events to the total weather data samples.
\item \textbf{Precision} measures the correctly classified extreme weather events, in the proportion of the total classified extreme weather events.
\item \textbf{Recall or detection rate} measures the proportion of correctly classified extreme weather events by the models, in the actual extreme weather events, to measure the ability for detecting extreme weather events.
\item \textbf{F1 score} measures the harmonic mean of the precision and recall.
\end{itemize}

It is expected that the models capture as many extreme events as possible, so as to effectively save lives and reduce damages. Therefore, the metrics above, accuracy, precision and recall, are considered as crucial ones when evaluating model performance. 

\section{Results}

The performance results of the experiments are as follows.

\begin{table*}[t]
\caption{Precipitation Model Performance \\ (SEM - Standard Error of the Mean, calculated by taking the standard deviation and dividing it by the square root of the sample size).}
\label{climatenets-result-table}
\vskip 0.15in
\begin{center}
\begin{small}
\begin{sc}
\begin{tabular}{lcccccr}
\toprule
Models & Loss & Accuracy & Precision & Recall & AUC & F1\_Score \\
\midrule
ConvNet    			& \begin{tabular}{c}0.4177\\$\pm 0.0259$\end{tabular} & \begin{tabular}{c}0.8671\\$\pm 0.0030$\end{tabular} & \begin{tabular}{c}0.2613\\$\pm 0.0050$\end{tabular} & \begin{tabular}{c}0.9060\\$\pm 0.0058$\end{tabular} & \begin{tabular}{c}0.9533\\$\pm 0.0014$\end{tabular} & \begin{tabular}{c}0.8669\\$\pm 0.0030$\end{tabular} \\\hline
SAConvNet 			& \begin{tabular}{c}0.2357\\$\pm 0.0127$\end{tabular} & \begin{tabular}{c}0.9052\\$\pm 0074$\end{tabular} & \begin{tabular}{c}0.3384\\$\pm 0139$\end{tabular} & \begin{tabular}{c}\textbf{0.9369}\\$\pm 0.0098$\end{tabular} & \begin{tabular}{c}0.9678\\$\pm 0.0021$\end{tabular} & \begin{tabular}{c}0.9048\\$\pm 0.0074$\end{tabular} \\\hline
SACOnvNet + Highway	& \begin{tabular}{c}0.0.0876\\$\pm 0.0079$\end{tabular} & \begin{tabular}{c}\textbf{0.9688}\\ $\pm 0.0034$\end{tabular} & \begin{tabular}{c}\textbf{0.6423}\\$\pm 0.0308$\end{tabular} & \begin{tabular}{c}0.8830\\$\pm 0.0061$\end{tabular} & \begin{tabular}{c}0.9763\\$\pm 0.0011$\end{tabular} & \begin{tabular}{c}0.9693\\$\pm 0.0032$\end{tabular} \\
\bottomrule
\end{tabular}
\end{sc}
\end{small}
\end{center}
\vskip -0.1in
\end{table*}

As is clearly shown in the table \ref{climatenets-result-table}, the proposed Self-Attention CNN model with the Highway networks achieves the best overall accuracy of 97\% across both classifications (extreme precipitation vs. non-extreme precipitation), a 12\% improvement as compared to the classical CNN model result by \cite{Davenport21}. It is capable of accurately identifying more than 88\% of severe precipitation days as extreme precipitation days (EPD), comparable to the classical CNN model. Extreme precipitation occurred on 64\% of days classified as EPD patterns, 150\% better than the classical CNN model result. And, only fewer than 12\% of days, classified as non-EPD patterns, resulted in severe precipitation. 

\begin{figure*}
	\centering
	\subfigure[ConvNet Confusion Matrix]{\includegraphics[width = 3in]{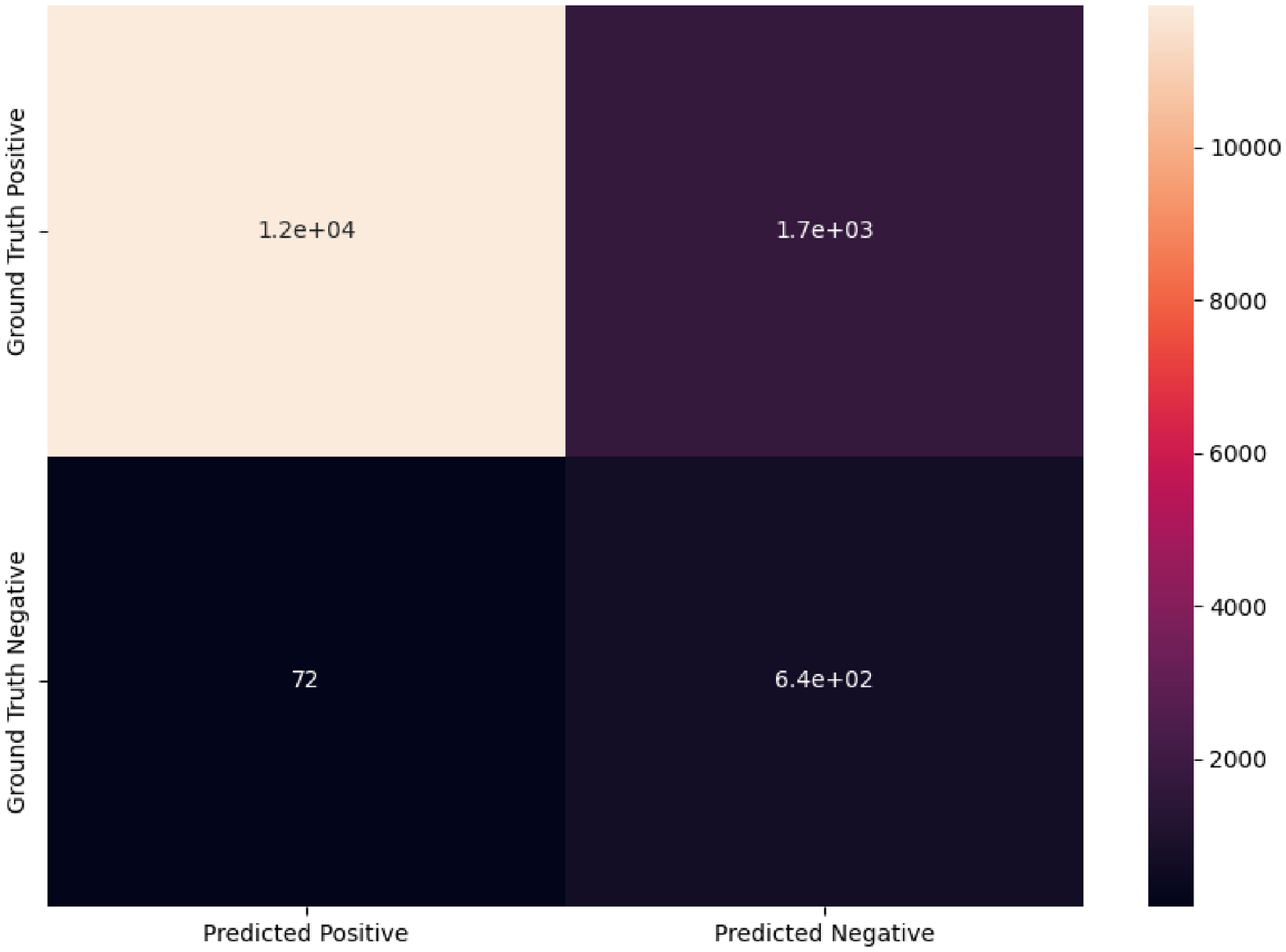}} 
	\subfigure[SAConvNet Confusion Matrix]{\includegraphics[width = 3in]{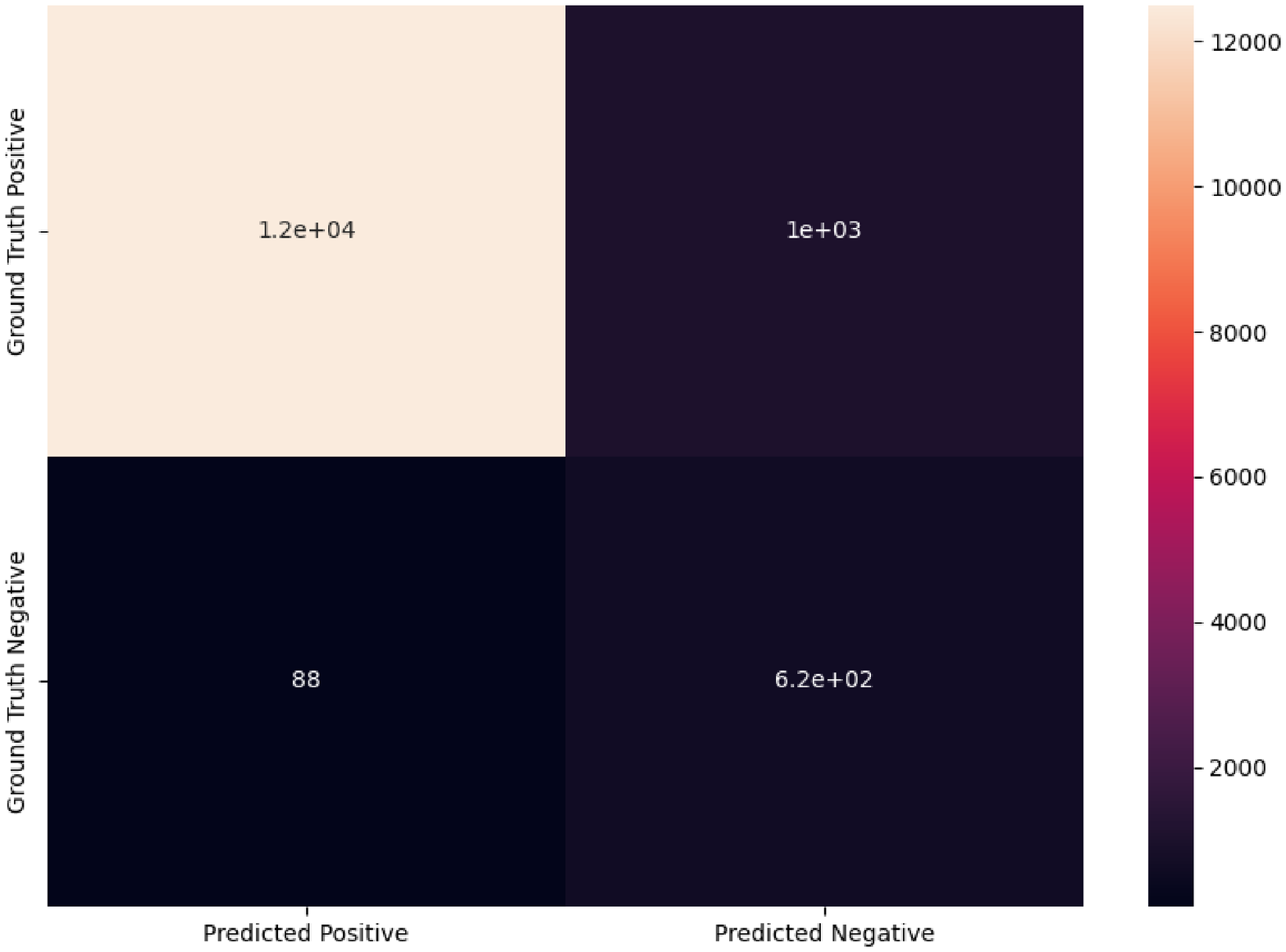}}
	\subfigure[SAConvNet + Highway Confusion Matrix]{\includegraphics[width = 3in]{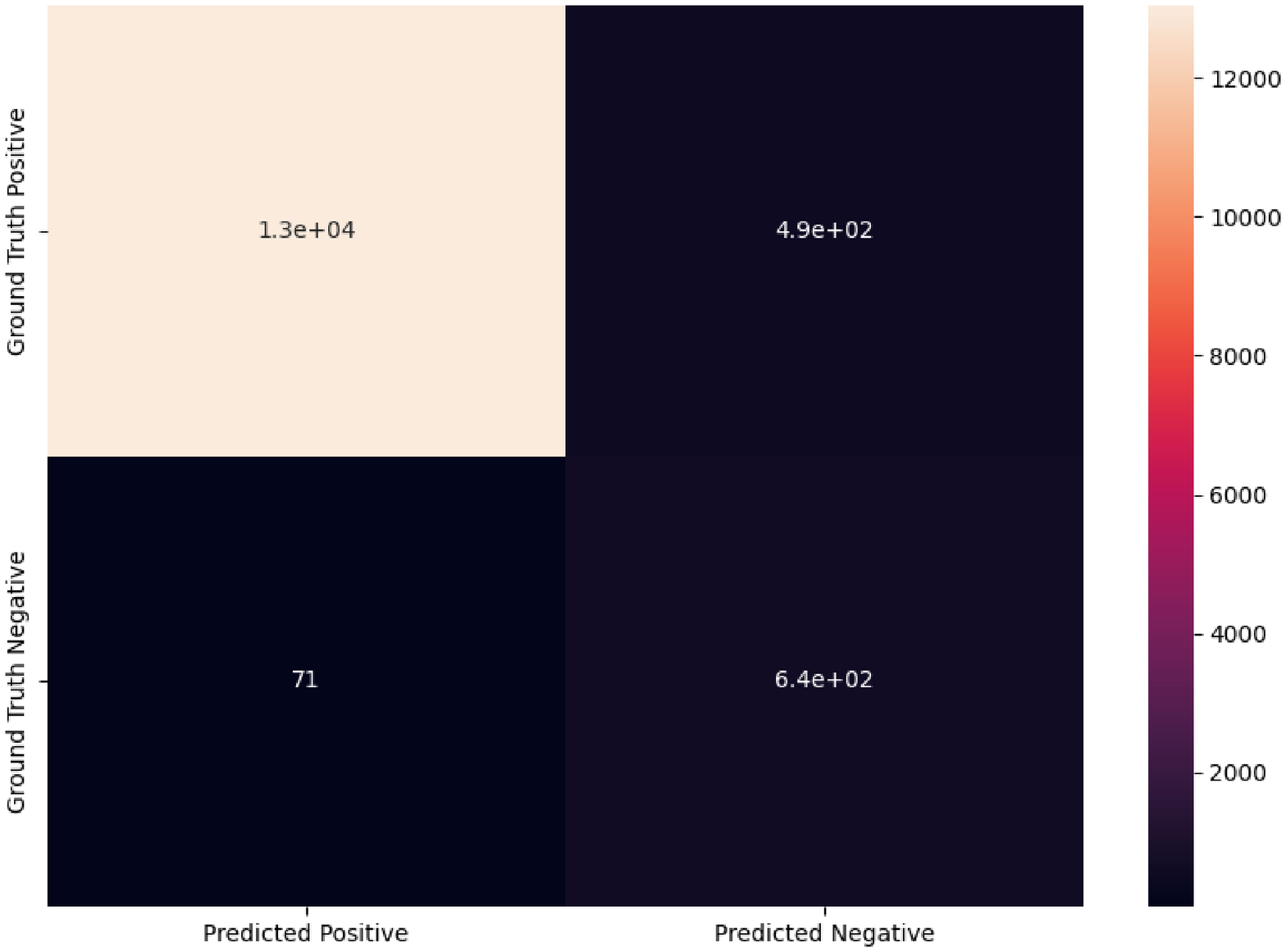}}
	\caption{Confusion Matrices}
	\label{confusion-matrices}
\end{figure*}

Without the Highway network, the proposed Self Attention CNN model achieves even better performance, nearly 94\% recall rate, in successfully identifying severe precipitation days as EPD, with only as low as 6\% of EPD classified as non-EPD. The overall accuracy of SAConvNet is still above 90\%, which is still well above the classical CNN model results by 4\%. The confusion matrices of our experiments are shown in Figure \ref{confusion-matrices}.

\begin{figure*}
\vskip 0.2in
\begin{center}
\centerline{\includegraphics[width=0.75\textwidth]{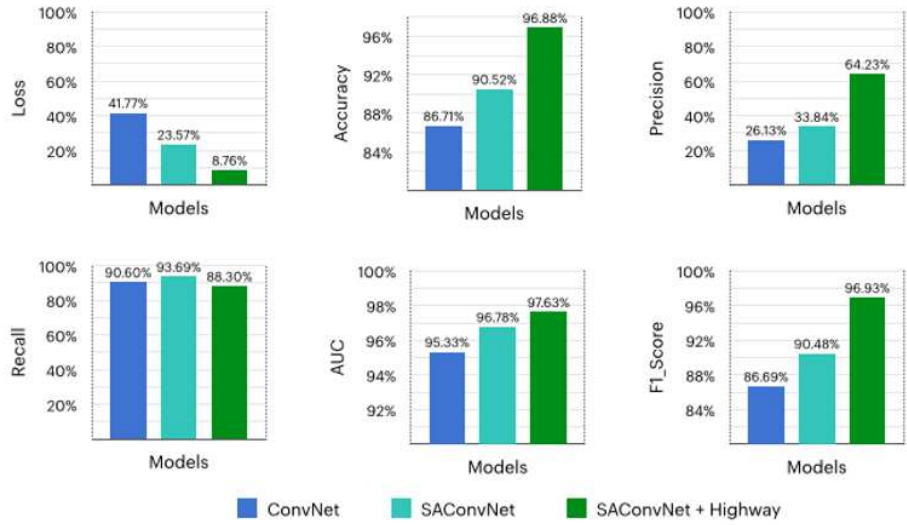}}
\caption{Precipitation Model Performance.}
\label{climatenets-result-chart}
\end{center}
\vskip -0.2in
\end{figure*}

\section{Applications}

The offered solutions have apparent commercial benefit. Extreme precipitation occurrences have caused substantial (and will continue to cause) economic damage and loss of life. Deep Learning techniques like these might undoubtedly aid in mitigating the losses caused by climate change and catastrophic weather occurrences.

Because the models are effective and simple, the proposed solution can be deployed at a low cost to weather stations on a PC server with a GPU card to assist in the prediction and classification of extreme weather events, providing short-term warnings to individuals in a designated area, planning for evacuation protocols, and estimating the severity of an extreme weather event if it occurs. The total solution has yet to be evaluated in a real-world setting, but it serves as another potential future enhancement to this study.

The models can also be used to enhance research in domains such as climate change. Climate scientists may be able to uncover additional answers to concerns about the consequences of climate change by linking extreme weather occurrences to climate variables in the data, and they may even be able to relate the effects to the precise causes of climate change. While the models have yet to be fully tested in real-world settings, they have the potential to be a tool for climatologists to anticipate catastrophic weather occurrences and climate change far into the future, as previously stated.

\section{Conclusions and Future Works}

In this study, we propose a unique Self-Attention Convolutional Neural Network and its variants, and evaluate these models using real-world datasets, with the goal of harnessing breakthroughs in AI technology to aid in the struggle against extreme events caused by accelerating climate change.

During the past decades, it has been well acknowledged that the weather intensity has increased significantly when extreme weather events occur. This subsequently has an increasing influence on the fraction of extreme precipitation days above the p95 threshold. Extreme precipitation intensity has increased in tandem with preferred increases in moisture flow. These findings show that thermodynamically driven changes in severe precipitation arise earlier than dynamically driven changes. The proposed models are very effective at distinguishing between extreme and non-extreme weather days.

On average, the proposed models are capable of identifying extreme weather days based mostly on climate variables. These variables are consistent with the pattern of high moisture flux. Thus, despite not having any external geographic information such as maps in the extreme precipitation timeseries, the proposed models uncover physically meaningful features in the regions.

On the other hand, some extremes might still be missed by the algorithms. This is because extreme precipitation is often controlled by localized processes not reflected in the regional-mean daily precipitation. Additional input variables that reflect smaller-scale meteorological processes (e.g., vertical velocity, wind shear, and convective available potential energy) could further improve the models' accuracy for forecasting-related applications.
The outputs of the models could likewise be adjusted to concentrate on a smaller region or to capture days with sub-regional variability.

Although the models show improved predictive accuracy across most tasks, the area on which we focused our extreme precipitation data may not completely reflect the model's predictive power. As addressed above, future work can include creating a better, possibly more scalable model to handle large amounts of data. Few-shot learning and meta-learning could also be used to significantly accelerate model convergence.

Additionally, by incorporating different types of information, such as geographical and climate data, the models could provide insight into different causes of extreme precipitation and extreme weather events.

Finally, because the models are trained on variables that are available globally, they could be readily retrained to analyse extreme precipitation and possibly even additional types of extreme weather in other regions, such as non-US regions. This approach could be useful in regions where the limitations of climate models in simulating precipitation processes lead to high uncertainty in future changes in extreme precipitation.

Although the initial case study is confined to small data sets, the results demonstrate that deep learning can provide critical insight into the physical processes underlying changes in climate extremes. Because it is generalizable to a range of extreme events and regions, it represents a promising tool for both scientific understanding and the planning and adaptation required to reduce vulnerability to current and future climate change.

\bibliographystyle{alpha}
\bibliography{climatenets}

\end{document}